\def\TOG{}
\documentclass[journal]{IEEEtran}
\pdfoutput=1

\usepackage{ifpdf}
\usepackage{cite}
\usepackage{graphicx}
\usepackage{amsmath}
\usepackage{algorithmic}
\usepackage[caption=false]{subfig}
\usepackage{url}
\usepackage{amssymb}
\usepackage{colortbl}
\usepackage{hyperref}

\newcommand{\new}[1]{#1}
\newcommand{\revii}[1]{#1}
\newcommand{\changed}[1]{#1}

\makeatletter
\def\ps@IEEEtitlepagestyle{
  \def\@oddfoot{\copyrightnotice}
  \def\@evenfoot{}
}
\def\copyrightnotice{
  {\footnotesize
  \begin{minipage}{\textwidth}
  \centering
   \footnotesize 2475-1502~\textcopyright~2018 IEEE. DOI: \href{https://doi.org/10.1109/TG.2018.2861759}{10.1109/TG.2018.2861759} Personal use is permitted, but republication/redistribution requires IEEE permission. See http://www.ieee.org/publications standards/publications/rights/index.html for more information.
  \end{minipage}
  }
}

\begin{document}   
\title{Exploration in NetHack With Secret Discovery}

\ifdefined\TOG
	\author{Jonathan~Campbell
	  and~Clark~Verbrugge
	  \thanks{School of Computer Science, McGill University, Montr\'eal, Canada.  email: \texttt{jonathan.campbell@mail.mcgill.ca}, \texttt{clump@cs.mcgill.ca}}}
\else
	\author{Jonathan~Campbell
	  and~Clark~Verbrugge\\
	  School of Computer Science, McGill University, Montr\'eal, Canada.\\ \texttt{jonathan.campbell@mail.mcgill.ca}, \texttt{clump@cs.mcgill.ca}}
\fi

\ifdefined\TOG
	\markboth{IEEE Transactions on Games}%
	{Campbell and Verbrugge: Exploration in NetHack With Secret Discovery}
\fi

\maketitle

\begin{abstract}
\textit{Roguelike} games generally feature exploration problems as a critical, yet often repetitive
element of gameplay.  Automated approaches, however, face challenges in terms of optimality, as well as
due to incomplete information, such as from the presence of secret doors.  This paper presents an algorithmic
approach to exploration of roguelike dungeon environments.  Our design aims to minimize exploration time,
balancing coverage and discovery of secret areas with resource cost. Our algorithm is based on the
concept of occupancy maps popular in robotics, adapted to encourage efficient discovery of secret access
points.  Through extensive experimentation on \textit{NetHack} maps we show that this technique is significantly more
efficient than simpler greedy approaches \new{and an existing automated player}.  We further investigate optimized parameterization for the
algorithm through a comprehensive data analysis. These results point towards better automation for
players as well as heuristics applicable to fully automated gameplay.

\end{abstract}

\ifdefined\TOG
\fi

\section{Introduction}


\ifdefined\TOG
	\IEEEPARstart{M}{any}
\else
	Many
\fi
 video games place emphasis on the idea of exploration of the unknown. In
\textit{roguelikes}, a popular subset of Role-Playing Games (RPGs), exploration of the game space
is a key game mechanic, essential to resource acquisition and game progress.  The high level of
repetition involved, however, makes automation of the exploration process useful, as an assistance in game design,
for relieving player tedium in relatively safe levels or under casual play, and to ease control requirements for those operating
with reduced interfaces~\cite{rps-17-roguelikes}. Basic forms of automated exploration are found in
several roguelikes, including the popular \textit{Dungeon Crawl Stone Soup}.

Algorithmic approaches to exploration typically aim at being exhaustive.  
Even with full information, however, ensuring complete coverage can result in significant
inefficiency, with coverage improvement coming at greater cost as exploration
continues~\cite{chowdhury-16-exhaustive}.  Diminishing returns are further magnified in the presence of
``secret rooms,'' areas which must be intentionally searched for at additional, non-trivial resource cost,
and which are a common feature of roguelike games. 
In such contexts, the complexity is less driven by the need to be thorough, and more given by the need to
balance time spent exploring with respect to amount of benefit accrued (area revealed,
items collected).

In this work we present a novel algorithm for exploration of an initially unknown environment.
Our design aims to accommodate features common to roguelike games.  In particular, we aim for
an efficient, balanced approach to exploration, considering the cost of further exploration in relation to
the potential benefit.  We factor in the relative importance of different areas, focusing on
room coverage versus full/corridor coverage, and address the existence of secret rooms (secret
doors) as well.  Our design is inspired by a variation of occupancy maps, adapted from robotics into
video games~\cite{isla-13-occpres}.  In this way we can control how the space is explored, following
a probability gradient that flows from places of higher potential benefit.

Using the canonical roguelike \textit{NetHack} as environment, we compare this approach with a simpler, greedy algorithm more typical of a basic automated strategy, \new{as well as an existing NetHack bot, \textit{BotHack}, and optimal solution.}  NetHack gives us a realistic and
frequently mimicked game context, with uneven exploration potential (rooms versus corridors), critical
resource limitations (every move consumes scarce food resources), and a non-trivial, dungeon-like map
environment, including randomized placement and reveal of secret doors.
Compared to the greedy approach \new{and BotHack}, our algorithm shows improvement in overall efficiency, particularly
with regard to discovery of secret areas. We enhance this investigation with a deep consideration of
the many different parameterizations possible, showing the relative impact of a wide variety of
algorithm design choices.

Our design is intended to provide a core system useful in higher level approaches to computing game solutions,
as well as in helping good game design. For the former we hope to reduce the burden of exploration itself as a 
concern in research into techniques that fully automate gameplay. 

Specific contributions of this work include:
\begin{itemize}
	\item We heavily adapt a known variation of occupancy maps to the task of performing efficient
      exploration of dungeon-like environments \revii{that do not contain secret areas}.
    \item We further extend the exploration algorithm to address discovery of secret areas.  Locating
      and stochastically revealing an unknown set of hidden areas adds notable complexity and cost to
      optimizing such an algorithm.
    \item Our design is backed by extensive experimental work, validating the approach and comparing it
      with a simpler, greedy design \new{and an existing automated player}, as well as exploring the impact of the variety of different parameterizations
      available in our approach.
\end{itemize}

This work builds on a previous short (poster) publication, wherein we described the basic exploration algorithm \cite{campbell-17-exploration}.  Here we significantly extend that work, incorporating discovery of secret areas into the greedy and occupancy map algorithms and performing additional experimental comparison in that context. We also add a non-trivial regression analysis to better understand the importance of the many parameters involved in the algorithm design, as well as \new{adding a comparison to a well-known NetHack bot and to an improved optimal solution}.

\section{Related Work}
\label{sec:related}
	

Automated exploration or mapping of an environment has been studied in several fields, primarily including robotics, and is also related to the problems of coverage and graph traversal.

Exploration in robotics branches into many different subtopics, differentiated by the type of environment to be explored, amount of prior knowledge about the environment, and accuracy of robotic sensors. One frequently-discussed approach is simultaneous localization and mapping (SLAM), where a mobile robot must map a space while keeping precise its current position inside said space. Our game-based environment has a top-down view and thus position is always known, so we can avoid this issue. Julia et al.~\cite{julia-12-comparison} and Lavalle~\cite{lavalle-06-planning} provide good surveys of robotic exploration algorithms, with coverage on SLAM. \new{Thrun~\cite{thrun-02-mapping} offers a more general survey of robotic mapping with coverage of exploration.}

One method popular in robotics for exploring unknown environments is known as occupancy mapping~\cite{moravec-85-maps, moravec-88-sensor}. This approach, used in conjunction with a robot and planning algorithm, maps out an initially unknown space by maintaining a grid of cells over the space, with each cell representing the probability that the corresponding area is occupied (by an obstacle/wall, e.g.). With this data structure, knowledge within a certain confidence margin can be established about which areas of the space are traversable, with data from different sensors being combined to even out sensor inaccuracies.

This sort of representation of the observed map must then be leveraged to decide where to move next for efficient exploration. Strategies typically involve an ordering or choice of frontiers to visit, sometimes determined by an evaluation function which takes into account objectives like minimizing distance travelled or exploring the largest amount of map the fastest. Yamauchi described a strategy using occupancy maps to always move towards the closest frontier in order to explore a space~\cite{yamauchi-97-frontier}, with a focus on how to detect frontiers in imprecise occupancy maps. Gonz\`{a}lez-Ba\~{n}os and Latombe discuss taking into account both distance to a frontier and the `utility' of that frontier (a measure of the unexplored area potentially visible when at that position)~\cite{gonzalesbanos-02-indoor}, also taking into account robotic sensor issues. We use a similar cost-utility strategy for our evaluation function (with utility determined by probabilities in the occupancy map, as described later). Juli\'{a} et al. showed that a cost-utility method for frontier evaluation explores more of the map faster than the closest frontier approach, but in the end takes longer to explore the entire map than the latter since it must backtrack to explore areas of low utility~\cite{julia-12-comparison}. We aim to not visit areas of low utility, so this downside will not apply. \new{Amigoni~\cite{amigoni-08-evaluation} presents further discussion and comparison of frontier evaluation functions.}

\new{The exploration problem in robotics is also related to the coverage path planning problem, where a robot must compute a path that traverses the entirety of a known space. A cellular decomposition of the space is used in many such approaches.} For example, Xu et al. presented an algorithm to guarantee complete coverage of a known environment (containing obstacles) while minimizing distance travelled based on the boustrophedon cellular decomposition method, which decomposes a space into slices~\cite{xu-2014-aerial}. \new{Choset~\cite{choset-01-coverage} provides a comprehensive discussion and survey of selected coverage approaches. Pure algorithmic methods for coverage are related to the traveling salesman problem (TSP) or shortest Hamiltonian path problem. These algorithms need full knowledge of the environment at start, and are thus not applicable to NetHack.}

There have also been formulations of exploration as a graph traversal problem. Kalyanasundarum and Pruhs describe the `online TSP' problem as exploring an unknown weighted graph, visiting all vertices while minimizing total cost, and presented an algorithm to do so efficiently~\cite{kalyanasundaram-94-tours}. Koenig et al. analyzed a greedy approach to explore an unknown graph (to always move to the closest frontier), and showed that the upper bound for worst-case travel distances for full map exploration is reasonably small~\cite{koenig-01-greedy, tovey-03-greedy2}. \new{Hsu and Hwang demonstrate a provably complete graph-based algorithm for autonomous exploration of an indoor environment~\cite{hsu-98-indoor}.}

Research into exploration has also been done in the context of video games. Chowdhury and Verbrugge looked at approaches to compute a tour of a known environment for exhaustive exploration strategies for non-player characters in video games~\cite{chowdhury-16-exhaustive}. Baier et al. proposed an algorithm to guide an agent through both known and partially-known terrain to catch a moving target in video games~\cite{baier-14-prey}. \new{By contrast, our goal is to find static areas (unvisited and possibly secret rooms), not moving targets.} Hagelb\"{a}ck and Johansson explored the use of potential fields to discover unvisited portions of a real-time strategy game map with the goal of creating better game AI~\cite{hagelback-08-fow}. Our work, in contrast, focuses on \changed{constrained} exploration in sparse, dungeon-like environments, where exhaustive approaches compete with critical resource efficiency.

\section{Background}
\label{sec:background}


Three concepts underpin our work and will be briefly discussed below: the particular flavour of occupancy maps used as the basis for our exploration algorithm; the game used for our research environment; and a short elucidation on the concept of secret rooms and their presence in video games.

\subsection*{Occupancy Maps in Games}
Using the aforementioned occupancy maps from robotics as inspiration, Dami\'{a}n Isla created an algorithm geared towards searching for a moving target in a video game context~\cite{isla-05-occmap}. The algorithm has been used in at least one game to date~\cite{isla-13-occpres}.

Like the original occupancy map, a discrete grid of probabilities is maintained over a space (e.g., game map), but here a probability represents confidence in the corresponding area containing the target or not, instead of simply relating to traversability. A non-player character (NPC) can then use said map to determine where best to move in order to locate the target (e.g., the player).

\revii{The map is updated as follows. The NPC first observes if the player is within its current field-of-view (FOV) and the corresponding cells are set accordingly (1 for cell(s) containing the player, and 0 otherwise). The NPC will then begin moving towards the cell with the highest probability. If the target is in sight, the NPC can move directly towards them; otherwise, it can move towards the cell with the highest probability. All probabilities in the map then diffuse to their neighbours. Through diffusion, cells outside the FOV will contain a probability value based on their closeness to the last known player position, with a probability gradient flowing from that last known spot into adjacent areas which the player may have moved to.}

Diffusion for each cell $n$ outside the NPC's FOV at time $t$ is performed as follows (assuming each cell has 4 neighbours):

$P_{t}\left ( n \right ) = \left ( 1 - \lambda  \right ) P_{t-1}\left ( n \right ) + \frac{\lambda }{4}\sum_{n' \in 
\textit{neighbours}\left ( n \right )}^{ } P_{t-1}\left ( n' \right )$

\noindent{}where $\lambda \in [0, 1]$ controls the amount of diffusion.
 
Our implementation of occupancy maps borrows concepts from Isla's formulation, namely the idea of diffusion, which is repurposed for an exploration context.

\subsection*{NetHack}
NetHack is a popular roguelike video game created in 1987 and is used as the environment for our experiments. Gameplay occurs on a two-dimensional text-based grid of size 80x20, within which a player can move around, collect items, fight monsters, and travel deeper into the Mazes of Menace. To win the game, a player must travel through over 50 levels of the dungeon, fight the high priest of Moloch and collect the Amulet of Yendor, then travel back up through all the levels while being pursued by an angry Wizard and finally ascend through the five elemental planes~\cite{nethack-16-strategy}.

\begin{figure}[htpb]
	\centering
	\includegraphics[width=0.47\textwidth]{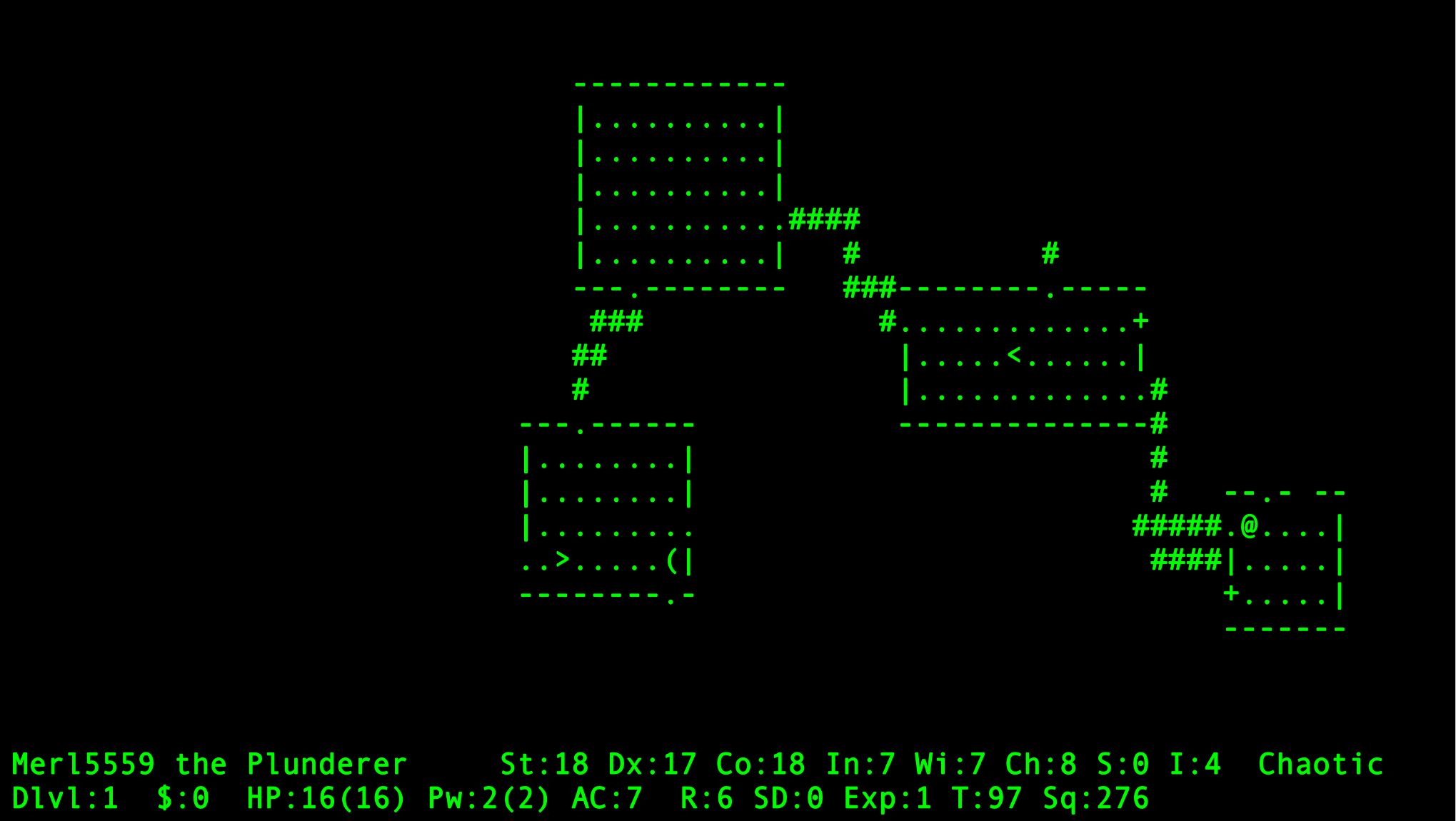}
	\caption{\small A game of NetHack where the player (`@' character, currently in the lower-right room) has explored about half of the level. A typical NetHack map is composed of corridors (`\#') that connect rectangular rooms. Room spaces (`.') are surrounded by walls (`$|$' and `-'), and doors (`+', `.'), which could lead to other, unvisited rooms, be dead-ends or circle around to already visited rooms.}
	\label{fig:nethack}
\end{figure}

Levels in NetHack consist of large, rectangular rooms (7-8 on average) connected by maze-like corridors. Levels can be sparse, with many empty (non-traversable) tiles. For the most part, levels are created using a procedural content generator, an advantage for conducting research in exploration since an algorithm can be tested on many different map configurations. At the start of each level, the player can see only their current room with the rest of the map hidden, and must explore to uncover more. An example of a typical NetHack map is shown in Figure~\ref{fig:nethack}; other maps can be seen in Figures~\ref{fig:nh_obvious_secret} and~\ref{fig:secmapvis1}.

Although map exploration is important, it is also exigent to do so in a minimal fashion. Movement in NetHack is turn-based (each move taking one turn), and the more turns made, the more hungry one becomes. Hunger can be satiated by food, which is randomly and sparingly placed within the rooms of a level, as well as being obtainable through monster combat~\cite{nethack-16-food}. Most food does not regenerate after having been picked up, so a player must move to new levels at a relatively brisk pace to maintain supplies. A player that does not eat for an extended period will eventually starve to death and lose the game~\cite{nethack-15-starvation}.

In this context, it is critical to minimize the number of actions taken to explore a level so that food resources are preserved. Rooms are vital to visit since they may contain food and items that increase player survivability, as well as the exit to the next level (needed to advance further in the game). Conversely, the corridors that connect rooms have no intrinsic value. Some may lead to dead-ends or circle around to already visited rooms. Exploring all corridors of a level is typically considered a waste of valuable actions. Therefore, a good exploration strategy will minimize visitation of corridors while maximizing the number of rooms visited.

\new{Several autonomous players in various states of maintenance currently exist for NetHack. Currently, only \textit{BotHack} has successfully completed the game~\cite{krajicek-15-bothack}. Another popular bot is \textit{TAEB (The Amulet Extraction Bot)}, which features a modular interface to easily support a variety of automatic players~\cite{taeb-13-repo}.}

\subsection*{Secret areas}
Secret areas are a popular element of game levels and motivate comprehensive exploration of a space. These areas are not immediately observable but must be discovered through extra action on the player's part. Secret areas can be a mechanism to reward players for thorough exploration, sometimes containing valuable rewards~\cite{hullett-10-fpsdesign}. In certain genres, they can also confer a sense of achievement for the player clever enough to find them. Gaydos \& Squire found that hidden areas in the context of educational games are memorable moments for players and generate discussion amongst them~\cite{gaydos-12-scientific}. Secret areas are perhaps most prevalent in roguelikes, with the prototypical roguelike games (Rogue, NetHack, et al.) all including procedurally-generated secret areas. \revii{These procedurally-generated secret areas differ from the ones discussed by Gaydos \& Squire since it could be mandatory to find them (if they happen to contain the level exit), and it can become repetitive and costly to continue attempting to discover them.}

Not much work has been done in terms of algorithms to search for secret areas. In terms of NetHack specifically, BotHack employs a simple secret area detection strategy. If either the exit to the next level has not yet been found or there is a large rectangular chunk of the level that is unexplored and has no neighbouring frontiers, it will start searching at positions that face that area~\cite{krajicek-15-bothack, krajicek-15-bothackthesis}.

\subsubsection*{NetHack implementation}
Secret areas in NetHack are created during level generation by marking certain traversable spots of the map as hidden. Both corridors as well as doors (areas that transition between rooms and corridors) can be marked as hidden (with a 1/8 chance for a door, and 1/100 chance for a corridor)~\cite{nethack-16-code}. On average, there are 7 hidden spots in a level. These hidden spots initially appear to the player as regular room walls (if generated as doors) or as empty spaces (if corridors) and cannot be traversed. The player can discover and make traversable a hidden spot by moving to a square adjacent to it and using the `search' action, which consumes one turn. The player may have to search multiple times since revealing the secret position is stochastic.

\begin{figure}[htpb]
	\centering
	\includegraphics[width=0.47\textwidth]{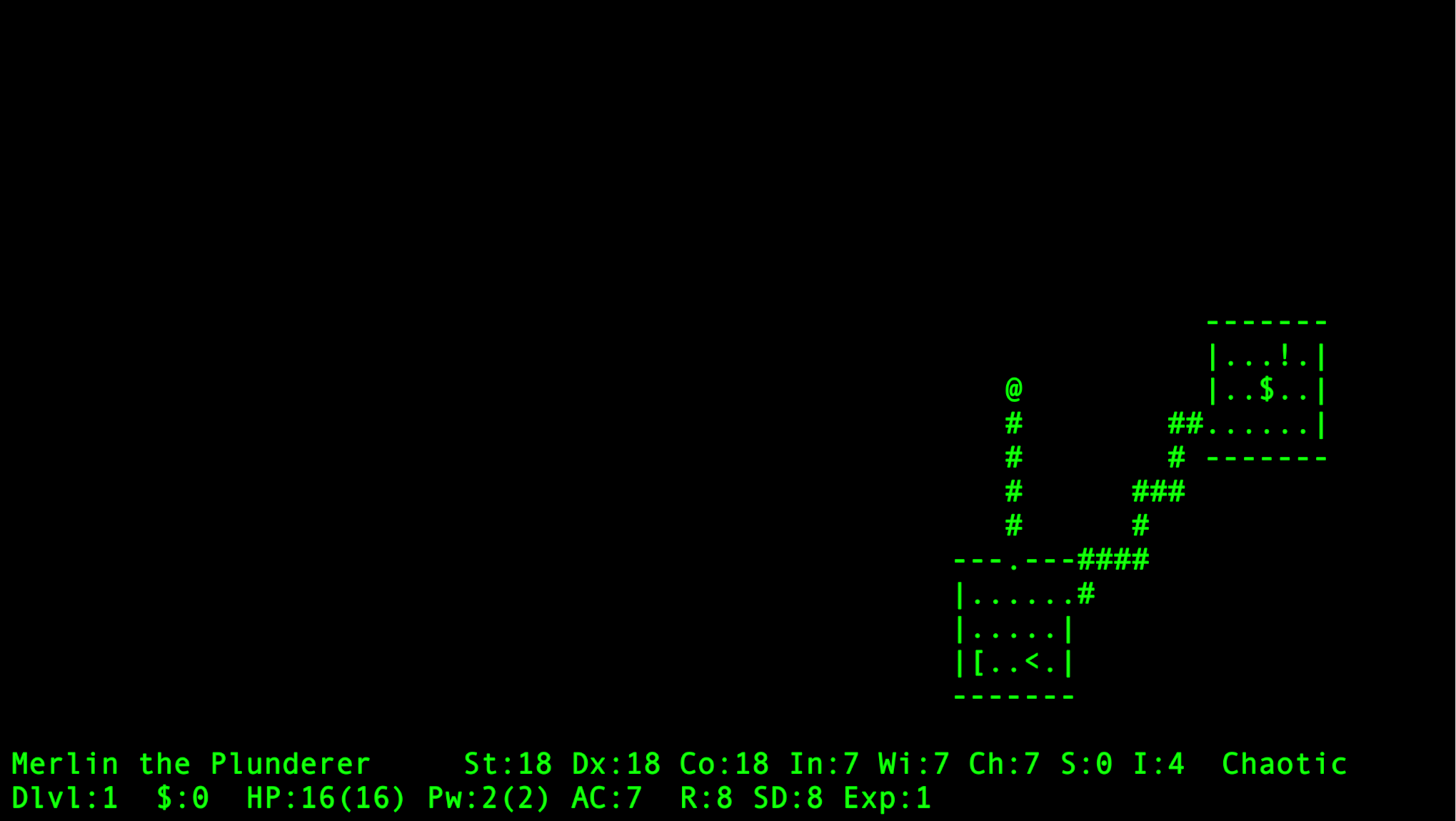}
	\caption{\small A NetHack map with the player having visited all non-secret positions. The vast majority of the map is still hidden, likely due to the presence of a secret corridor immediately north of the player's current position.}
	\label{fig:nh_obvious_secret}
\end{figure}

\changed{Searching consumes actions and thus food just like regular movement.} Therefore, the number of searches as well as the choice of locations searched must be optimized to preserve food resources. Intuitively, we would like to search walls adjacent to large, unexplored areas of the map, for which there do not appear to be any neighbouring frontiers. Similarly, corridors that end in dead-ends are also likely candidates for secret spots, as seen in Figure~\ref{fig:nh_obvious_secret}.

With the NetHack method of secret spot generation in mind, it becomes clear that it is not a good idea to attempt to discover every single hidden spot on a map. Some secret doors or corridors may lead nowhere at all, or perhaps lead to another secret door/corridor which leads to an area that the player has already visited. Depending on map configuration, the player may be able to easily spot such an occurrence and avoid wasting time searching in those areas. There is also a tradeoff between finding all secret (disconnected) rooms in a map and conserving turns; if only a small area of the map seems to contain a hidden area, spending a large effort to discover it may not be worthwhile.

\section{Exploration Approach}
\label{sec:metrics}


Below we detail the basic exploration algorithm involving occupancy maps, and contrast it with a simpler, greedy approach as well as an optimal solution. We begin by discussing the modified NetHack environment in which the algorithms will run, followed by an outline of each algorithm with and without support for detecting secret areas.

\subsection*{Environment}
A modified version of the base NetHack game is used to test our exploration algorithms. Mechanics that might confound results were removed, including monsters, weight limitations, locked doors, and certain dungeon features that introduce an irregular field of view. In addition, a switch to enable or disable generation of secret doors and corridors was added. Starvation was also removed, \new{but since we start each experiment with a regular hunger level, it is very unlikely for starvation to occur in the bounds of a full map exploration}.

The maps used in testing are those generated by NetHack for the first level of the game. The same level generation algorithm is used throughout a large part of the game, so using maps from only the first level does not limit generality. Later levels can contain special, fixed structures, but there is no inherent obstacle to running our algorithm on these structures; we are just mainly interested in applying exploration to the general level design (basic room/corridor structure).

The algorithms below use the standard NetHack player field of view. When a player enters a room, they are able to instantly see the entirety of the room, including its walls and doors. \new{In corridors, only the immediate neighbours of the current position can be observed. However, if the corridor continues in a straight line to an open room door, then the door and a small portion of the room past the door can be observed.}

\subsection*{Greedy algorithm \revii{with no secret areas}}
A greedy algorithm is used as baseline for our experiments, which simply always moves to the frontier closest to the player. This type of approach is often formalized as a graph exploration problem, where we start at a vertex $v$, learn the vertices adjacent to $v$, move to the closest unvisited vertex (using the shortest path) and repeat~\cite{koenig-01-greedy}. The algorithm terminates when no frontiers are left. We also take into account the particularities of the NetHack field of view as described above; when we enter a room, all positions in the room are set to visited, and its exits are added to the frontier list.

Note that this formulation will by nature uncover every traversable space on the map, both rooms and corridors alike.

\subsection*{Optimal solution}

For a lower bound on the average number of moves needed to visit all rooms on a NetHack map, we present the optimal solution. It will explore all rooms but not necessarily all corridors, similar to the occupancy map algorithm below, but must be given the full map in advance. \new{It will start in the centre of the initial room, and visit at least one door of each room on the map, guaranteeing visitation of all rooms.}

\new{We find the optimal path by treating it as an instance of a generalized shortest Hamiltonian path (GSHP) problem, which can then be reduced to a generalized Travelling Salesman Problem instance and run through a GTSP solver. The GSHP problem tries to find the shortest path in a graph that visits exactly (or at least) one vertex from each clustering of vertices.}

Our NetHack map is transformed into a graph as follows. Each room door will be represented as one vertex, and the set of doors of each room will form their own cluster. \revii{The center position of the player's starting room, which we term the `starting vertex',} will be placed by itself in another cluster. \revii{The cost of an edge is equal to the length of the optimal path between its vertices.} We then reduce this graph to a generalized Travelling Salesman Problem instance using the following method. Two additional `dummy' vertices are added to the graph, each in their own cluster. The first dummy vertex will have an edge to all other vertices with cost of 0, and the second will have an edge to only the \revii{starting vertex} and first dummy vertex with cost of 0. \new{This reduction was suggested by Lawler et al.~\cite{lawler-86-tsp} to translate a SHP problem to TSP, and we adapt it for the generalized version by placing each dummy vertex in their own cluster. We use \revii{the `GLNS' solver}~\cite{smith-17-glns} to obtain the solution to this GTSP problem.}

\subsection*{Occupancy maps \revii{with no secret areas}}

Our exploration strategy is described below. First, the use of occupancy maps as map representation will be described, followed by the application of diffusion and the frontier selection strategy.

As stated previously, the goal of the algorithm is to optimize exploration time by visiting as many rooms and as few corridors on a map as possible, without having foreknowledge of the map. In NetHack, rooms are much more likely to confer benefit to the player than corridors. Items, including food (necessary for survival), can spawn in rooms but not corridors; \new{further, corridors often contain many cycles and dead-ends. Excess exploration of a map then leads to food waste and more monster encounters (monsters spawn randomly every certain number of turns), increasing the probability of player death.}

\begin{figure}[htpb]
	\centering
	\includegraphics[width=0.47\textwidth]{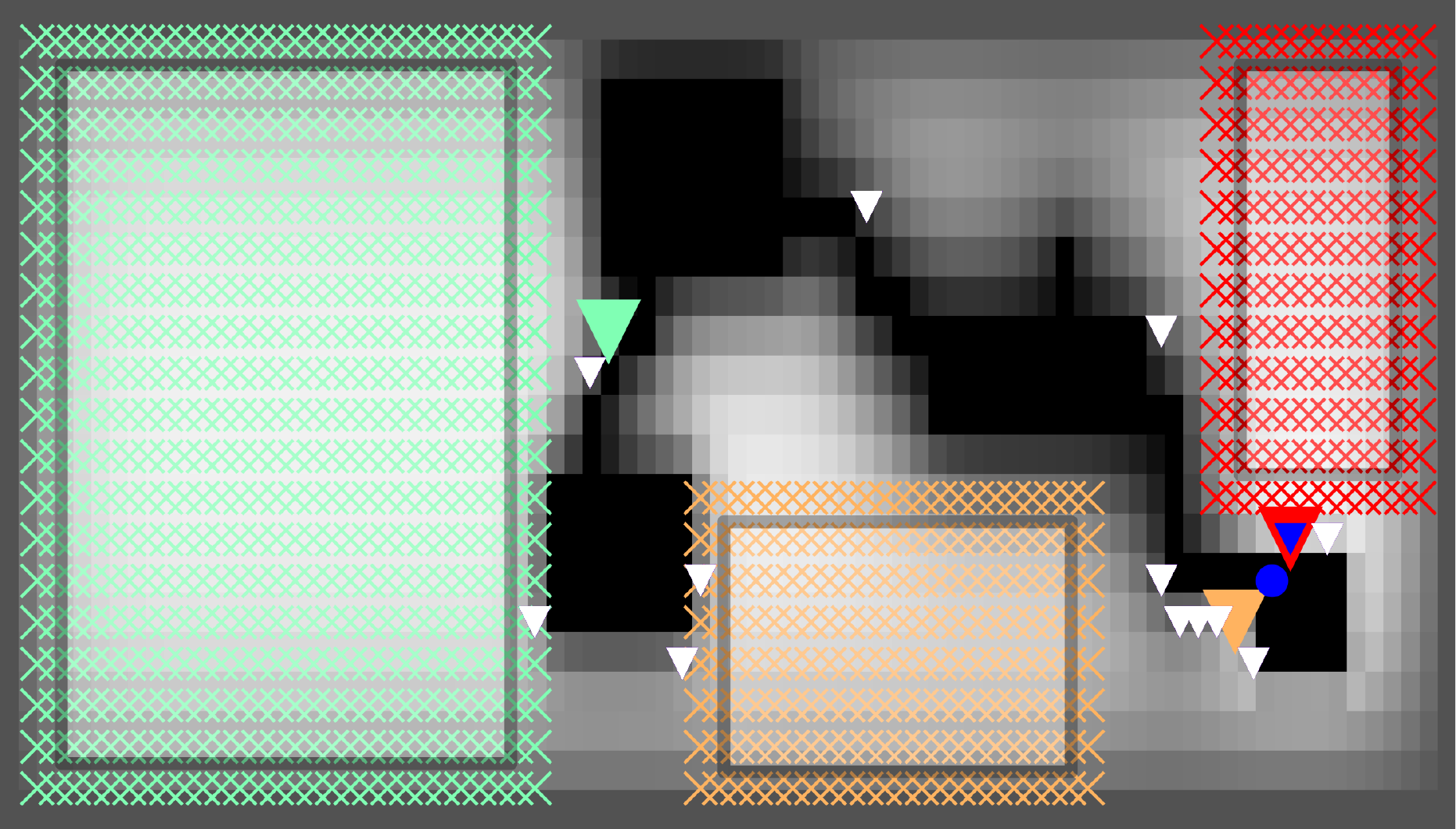}
	\caption{\small Visualization of an occupancy map corresponding to the NetHack level of Figure~\ref{fig:nethack}. Brighter areas are more likely to contain an undiscovered room. The player is shown as a circle in the lower-right room, and current target frontier shown as nearby inset triangle. Components are marked with a criss-cross pattern \revii{with rectangles inset}, and their associated frontiers are shown as large triangles. Other frontiers that will not be visited (due to being in areas of low probability and/or not having an associated component) are shown as small white triangles. Components without frontiers are unmarked.}
	\label{fig:occmapvis1}
\end{figure}

\subsubsection*{Representation}

To represent the map of a NetHack level, we use a structure similar to an occupancy map. However, instead of simply keeping track of obstacles, we instead assign probabilities to cells based on the likelihood that the corresponding position in the NetHack map is part of an unvisited room. If the corresponding position has already been visited, or observed to be a wall or other impassable tile, then the cell will have a probability of 0. Otherwise, it will have a non-zero value representing the confidence of it being part of an unvisited room, relative to all other cells in the map.

\changed{Initially, all probabilities in the occupancy map are set to the same value, equal to $1 / |total\_num\_cells|$. As the agent moves around, newly observed walls/impassable tiles will be set to 0, and the map is renormalized to ensure total probability sums to 1. Diffusion, discussed in the next section, also affects the probability values.}

\new{One parameter of the algorithm allows for the cells on the borders of the occupancy map to start with a fixed low probability.} This trick is done to discourage the agent from exploring frontiers near the borders of the map, since those often turn out to be dead-ends. As discussed further below, by lowering the probability of cells adjacent to a frontier, the algorithm will be less likely to visit them.

Figure~\ref{fig:occmapvis1} gives a visualization of a sample occupancy map, with brighter areas corresponding to higher probabilities.

\subsubsection*{Diffusion}

We here adapt the diffusion of probabilities from Isla's algorithm in order to better identify frontiers of low utility, optimizing our exploration time by ignoring such frontiers. Below, we discuss the reason for diffusion, as well as how and when it occurs.

\changed{By diffusing probabilities, the zero probability of observed space (visited rooms and corridors) will seep through the map.} Then, if a frontier is surrounded on all sides by low probabilities (to a certain depth of surroundings), we can easily identify that frontier as probably not leading to a new room, and thus ignore it. Figure~\ref{fig:occmapvis1} shows these low utility frontiers as small white triangles. \changed{This mechanism is a key part of the algorithm's efficiency, helping to decide which frontiers to visit and which to ignore. It also helps to determine which large, empty areas of the map could contain a new room, discussed further below.}

One detail of note is the minimum probability value for a frontier to be considered of low utility. This value is called the probability threshold. The threshold controls in a general sense the cutoff for exploration in areas of lower benefit; a higher value will mark more frontiers as unhelpful and thus focus exploration on areas of higher potential benefit (giving a tradeoff between time and amount explored). This threshold can be fixed throughout exploration, or in another formulation, vary depending on the percentage of map explored (ignoring more frontiers as more of the map is explored).

Diffusion is performed by imparting each cell with a fragment of the probabilities of its neighbouring cells, as given in the diffusion formula in section~\ref{sec:background}. Diffusion is only run when a new part of the map is observed (i.e., new room or corridor), so that probabilities will not change while travelling to a frontier through explored space. Further, cells corresponding to walls, impassable tiles or visited positions on the NetHack map are always set to 0 and cannot be changed during diffusion.

\subsubsection*{Frontier selection (planning)}

\changed{The main method to optimize exploration is to visit the frontiers that are most likely to lead to an as yet unseen room. Some frontiers can be outright rejected through the diffusion mechanism, but that still leaves many frontiers whose utility is harder to judge. To get a better idea of which frontiers are better than others and which should not be visited at all, we split the \new{unexplored areas of a} NetHack map into different rectangular areas, or \textit{components}, each with a certain probability of containing a room, and each being matched to a single frontier. In this way, the utility of a frontier can be based on the probabilities of its adjacent cells. Then, each component will be evaluated to choose which one to explore next. This procedure is detailed forthwith.}

Components are created by \revii{treating the occupancy map as a graph with edges between neighbouring cells of unexplored space}, that is, empty space (` ') on a NetHack map. Each disjoint set of unexplored cells is split into a series of maximal rectangles, and each rectangle is considered a component. To increase separation of components, edges are only created between cells that have more than a certain number of traversable neighbours, in order to eliminate narrow rectangular alleys of high probability cells. Cells are also ignored during edge creation if their probability value is above a certain threshold value (a parameter specified at start, which can be equal to or differ from the earlier frontier probability threshold).

Some components are ignored due to insufficient size or absence of neighbouring frontiers. If a component is smaller than the minimum size of a NetHack room, there is no point in trying to reach it. Likewise, if a component has no neighbouring frontiers, it cannot contain a room since there is no access point (unless secret doors/corridors are enabled, as discussed later). \new{A frontier is considered to neighbour a component if there is a straight line through unexplored space between it and the closest cell in the component.}

\new{One special case arises when the agent is in a corridor and has partially observed a room directly at the end of the corridor, as described in the Environment section. In this case, the cells corresponding to the observed room are put into their own component, not subject to any size restriction, and visitation of this component is prioritized over all others, since it is guaranteed to be a new room.}

The visualization of a sample occupancy map in Figure~\ref{fig:occmapvis1} gives an idea of the process, with three components marked using a criss-cross pattern. The unmarked area in the upper-middle is ignored since it has no neighbouring frontiers.

The list of valid components is then passed through an evaluation function to determine which best maximizes a combination of utility and distance values. Utility is calculated by summing the probabilities of all cells in the component, normalized by dividing by the sum of all probabilities in the map. \changed{Distance is calculated by taking the \revii{optimal} distance between the player and the neighbouring frontier closest to the component, normalized by dividing by the sum of player-frontier distances for each component.} With the normalized utility and distance values, we pick the component that maximizes $(1 - \alpha) \textit{norm\_prob} \: + \: \alpha \ast (1 - \textit{norm\_dist})$, where $\alpha$ controls the balance of the two criteria.

Once the best component is determined, the algorithm will move to its associated frontier. On arrival, it will learn new information about the game map, update the occupancy map, and run diffusion. Components will then be re-evaluated and a new frontier chosen. Exploration terminates when no valid components remain.

\subsection*{Greedy algorithm with secret rooms}

A trivial adaptation can be made to the basic greedy algorithm in order to search for secret areas. When entering a room, before proceeding to the next frontier, each wall of the room is searched for secret doors for a certain number of turns. \new{Walls that do not have at least three empty spaces adjacent to the space beyond them are ignored (since in such case it would be not be possible for a secret corridor to exist beyond).} Searches are also performed in dead-end corridors. If a secret door/corridor is discovered, it is added to the frontier list as usual. Exploration ends when no frontiers or search targets remain.

For efficiency, searching for secret doors in a room is done by first choosing the unsearched wall closest to the player, then moving to a spot adjacent to the wall that also touches the most walls still needing to be searched (since searching can be performed diagonally).

Note that this approach is not capable of finding all secret corridors in a level, since they may rarely appear in corridors other than dead-ends. However, searching all corridors would be too taxing to handle this rare occurrence. The below occupancy map approach also ignores these rare secret corridors.

\subsection*{Occupancy maps with secret rooms}

The occupancy map algorithm has a natural extension to support the discovery of secret door and corridor spots. In the original case, components of high probability in the occupancy map with no neighbouring frontiers would be ignored, but here, these components are precisely those that we would like to investigate for potential hidden doors/corridors. \new{We also immediately search at any dead-end corridor, due to the relative rarity of non-secret dead-ends.} Below we detail the adjustments necessary for this extension.

The first modification relates to the component evaluation function. Since these `hidden' components have by definition no neighbouring frontiers, \changed{there is no frontier to use when calculating distance to player. Instead, we calculate distance to a particular room wall or dead-end corridor adjacent to the hidden component.} \new{We also ignore hidden components whose area is below a specified minimum secret room size.}

The selection of such a wall or dead-end corridor for a hidden component requires its own evaluation function. This function will also consider both utility and distance. Utility is given by the number of searches already performed at that spot \changed{and distance taken as length of the \revii{optimal} path from the spot to the player.} Both search count and distance are normalized, the former by dividing by the sum of search counts for all walls, and the latter by dividing by the sum of distances for all walls. We then pick the spot that minimizes $(1 - \sigma) \textit{norm\_count} \: + \: \sigma \ast \textit{norm\_dist}$, where $\sigma$ is the parameter that controls the balance of the two criteria. The value is minimized to penalize larger distance and higher search counts.

Walls whose distance from the component exceed a specified maximum will be ignored, as well as walls that have already been searched over a specified maximum. \new{Further, like the greedy algorithm, walls must have at least three empty spaces adjacent to the space beyond them. For similar reasons, they must also have a straight line path through empty space to the closest component cell less than 10 units away.}

The selected wall/corridor spot is used in place of a frontier in component evaluation which proceeds as described earlier. If, after evaluation, a hidden component is selected, then we will move to the closest traversable spot adjacent to the component's associated wall/corridor spot. In case of ties in closest distance, the spot adjacent to the most walls will be chosen to break the tie, since searches performed at a position will search all adjacent spots (including diagonally).

When the player reaches the search location, the algorithm will use the search action for a specified number of turns before re-evaluating all components and potentially choosing a new search target or frontier to visit. If a secret door or corridor is discovered while searching, it is added to the open frontier list and its probability in the occupancy map is reset to the default value. If nothing is revealed after a certain number of searches (a parameterized value), then it will no longer be considered as a viable search target; it is possible for a hidden component not to contain a secret area.

Exploration terminates when no components are left, or only hidden components remain, none having viable search targets.

\begin{figure}[!htb]
\centering
\subfloat{%
  \includegraphics[width=0.47\textwidth]{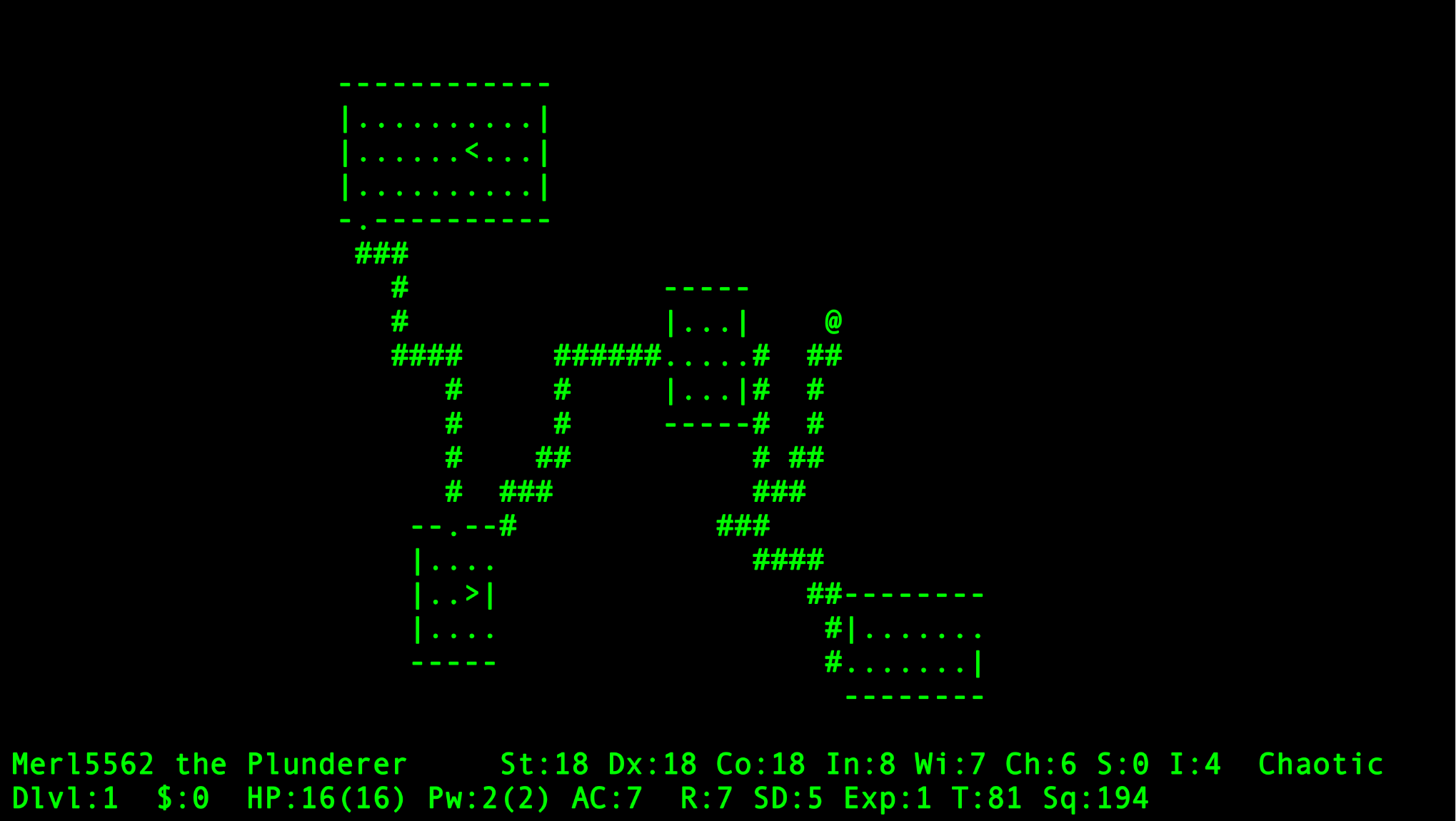}%
  }\par
\subfloat{%
  \includegraphics[width=0.47\textwidth]{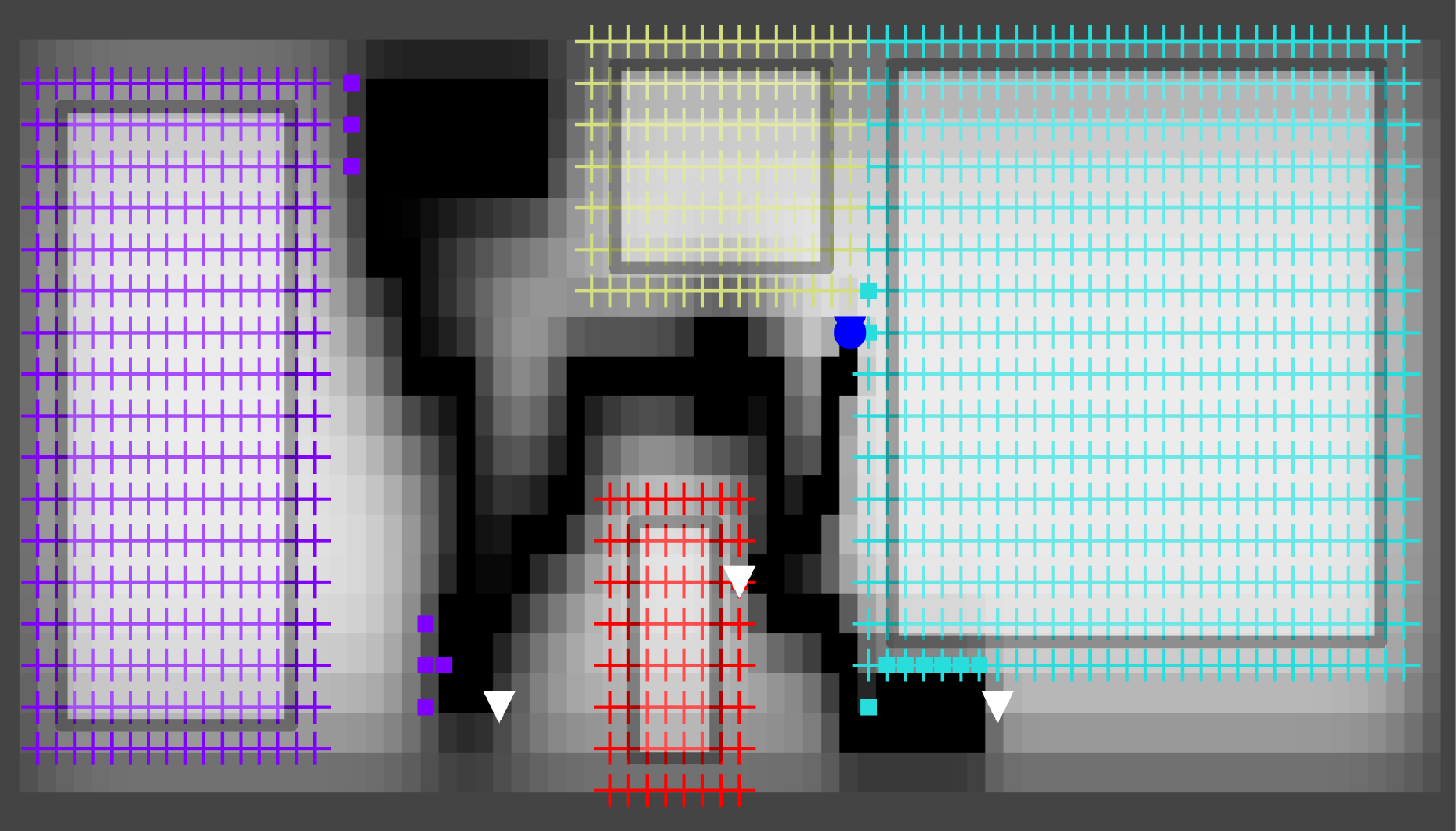}%
  }
\caption{Visualization of a sample occupancy map (bottom) and corresponding NetHack map (top) with secret doors/corridors enabled. Squares near each component represent the room walls and dead-end corridors that satisfy the distance and search count criteria. \revii{Components are identified by a plus-sign pattern with rectangles inset.}}
\label{fig:secmapvis1}
\end{figure}

Figure~\ref{fig:secmapvis1} presents a visualization of a sample occupancy map with secret doors/corridors enabled and corresponding NetHack map. The component on the left has no neighbouring frontiers and is thus considered a hidden component; nearby walls that will be considered for searching during evaluation are marked with squares. Meanwhile, the right-most component does have an open frontier (towards the lower-right), but the player is currently prioritizing the search of a dead-end corridor that may also open into that component.

\section{Experimental Results}
\label{sec:results}


Results will be shown below for the greedy and occupancy map algorithms as a function of the exhaustive nature of their searching, followed by results for the algorithms that can search for secret areas. We will look first at the metrics to be used for comparison of the algorithms.

\subsection*{Exploration metrics}
To evaluate the presented exploration strategies, we use as metrics the average number of actions per game (lower is better) as well as average percentage of rooms explored, taken over a number of test runs on randomized NetHack maps. As will be seen below, the presented algorithms tend to do quite well on these metrics. Thus, to get a more detailed view of map exploration which penalizes non-exhaustive exploration, we also use the percentage of maps in which all rooms were explored, which we term the `exhaustive' metric.

For algorithms that support detection of secret areas, two further metrics are used: the average percentage of secret doors and corridors found, and the average percentage of `secret rooms' found. Although these metrics are adequate for comparison purposes, neither are ideal, and it is important to understand limitations in evaluating secret room discovery.

The average percentage of secret doors/corridors found is problematic since it does not correlate well with actual benefit -- only a handful of secret spots will lead to undiscovered rooms and so be worth searching for. Further, it is biased towards the greedy algorithm, since that algorithm will search all walls, and so have a higher chance to discover more secret doors than the occupancy map algorithm, which will only search areas likely to contain secret rooms.

The average percentage of `secret rooms' found is also problematic \revii{due to its nebulous definition. In NetHack, rooms themselves are not marked as secret; marked instead are individual corridors and doors that may or may not lead to rooms disjoint from the non-secret part of the map. In this context, we can only define `secret rooms' to be} any room not directly reachable from the player's initial position in the level. However, this definition thus makes the metric dependent on map configuration: a map could exist such that the player actually starts in a `secret' room, separated from the rest of the map by a hidden door, and only that one door would have to be found for a full score on this metric to be given.

Further, while almost all maps tend to contain secret doors or corridors, only about half of all maps contain secret rooms as defined above (in the other half, secret doors/corridors still exist but do not lead anywhere). This discrepancy also skews the secret room metric since maps containing no secret rooms will still get a full score using that metric. \revii{We would in future like to look into better metrics for discovery of secret areas that address these issues.}

\subsubsection*{Food}

Although possible, we do not use amount of food collected as a metric. Food in the form of items is usually uniformly randomly distributed amongst rooms, and so is highly correlated with the percentage of rooms explored. \new{Further, this type of food is very scarce; in our experiments, less than half of all maps contained food, and those that did almost always had only one food item. Taking into account food pickup would thus not in any significant way alter the presented exploration results, since we already aim to visit all rooms, and deviating to pick up food in a room would only add 5-10 actions on average in half of the runs for all algorithms.}

\new{The other type of food comes from monsters; on death, monsters drop corpses, some of which can be eaten. However, inclusion of pickup of this food source would require monster combat to be integrated into our approach, which is a significant component on its own. We therefore choose to focus for now on the pure exploration method.}

\subsection*{Exhaustive approaches}

\begin{figure}
\centering
	\includegraphics[width=1\linewidth]{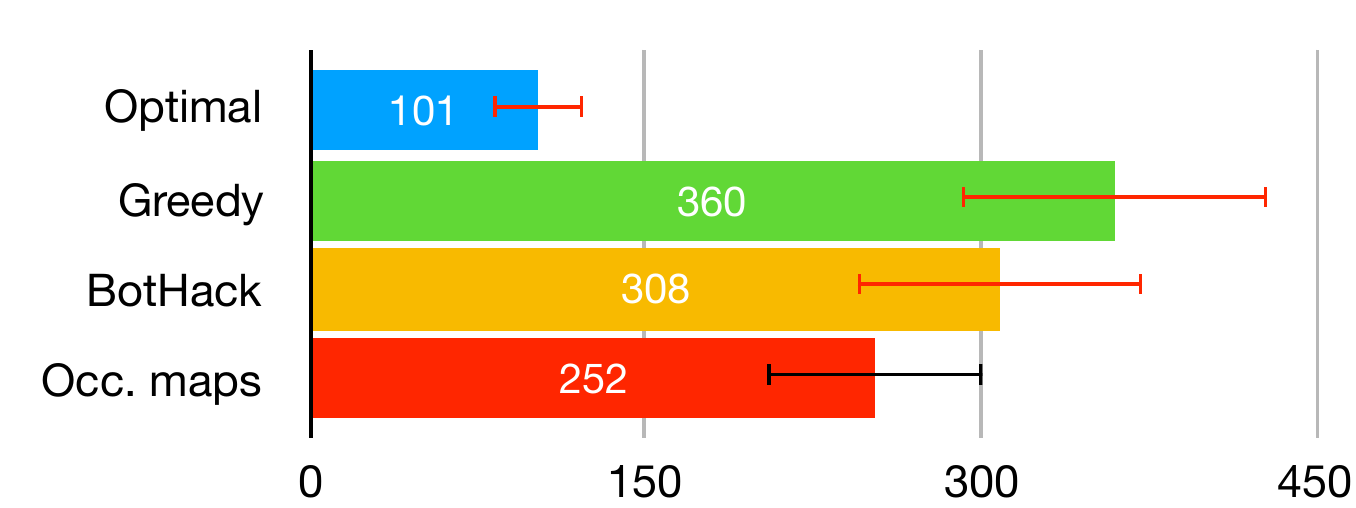}
	\caption{\small Average number of actions taken by the \new{optimal solution}, greedy algorithm, \new{BotHack}, and occupancy map algorithm for exhaustive room exploration with best performing parameters. The average over \changed{500} runs on different randomly-generated NetHack maps is taken, with standard deviation presented as error bars. \new{Note that the high standard deviation for all algorithms is due to the variance in NetHack map size and structure. P-value for occupancy map algorithm compared to greedy and BotHack \revii{is $<$ 0.01}, representing a significant difference.}}
	\label{fig:comp1}
\end{figure}

Figure~\ref{fig:comp1} presents results for the exhaustive exploration approaches (those that explore all rooms on a map). Each result is an average over \changed{500} runs on different randomly-generated NetHack maps. The greedy algorithm comes in at \changed{360} average actions per game, while the average for the fastest occupancy map model (with parameters that gave complete exploration on 98\% of all runs) is \changed{252} actions. \new{We also show the average actions for \textit{BotHack}, a general automated bot for NetHack, which comes in at 308 actions. To obtain the result, we modify BotHack to not search for secret doors/corridors, and terminate upon deciding to go to the exit of a level.}

The greedy algorithm is slowest since it by nature explores all corridors, while the occupancy map algorithm limits exploration to areas likely to contain rooms. The greedy algorithm is also a bit more reliable for complete room discovery than the occupancy map algorithm: we cited in the figure the occupancy map model that discovered all rooms in 98\% of runs, meaning that a small number of runs failed to discover all rooms on the map (missing one or two rooms in those cases).

In the same figure we present the result for the optimal solution for room visitation, which visits all rooms in \changed{101} actions on average. This approach can only be applied to a fully-known map, and so does not lend itself to exploration, but is instructive as a lower-bound.

\begin{figure}
\centering
	\includegraphics[width=1\linewidth]{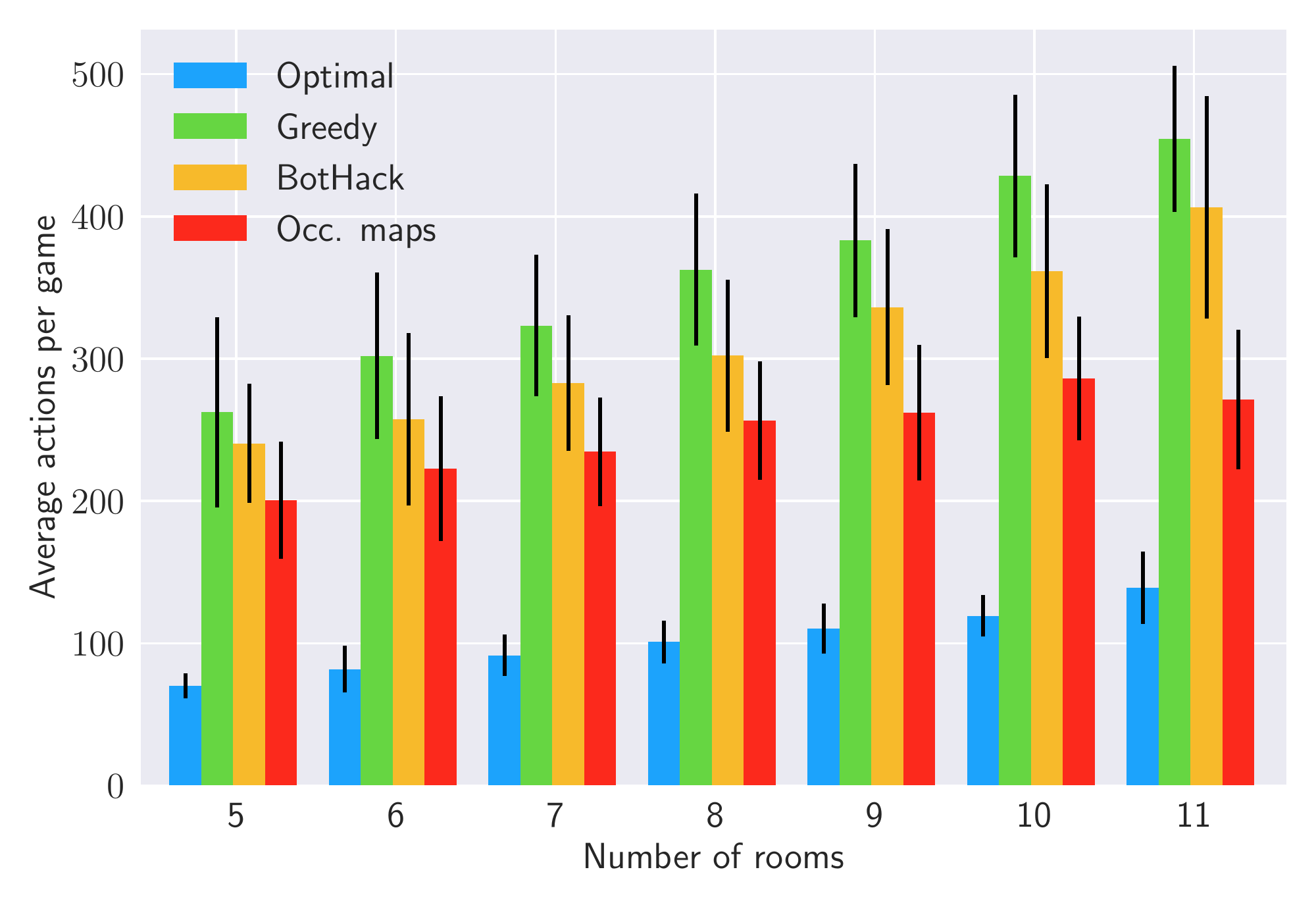}
	\caption{\small \new{Average number of actions taken with respect to number of rooms per map by the optimal solution, greedy algorithm, BotHack, and occupancy map algorithm for exhaustive room exploration with best performing parameters. The average over 500 runs broken down per number of map rooms is taken, with standard deviation presented as error bars.}}
	\label{fig:comp1b}
\end{figure}

\new{To further break down the results of Figure~\ref{fig:comp1} and compensate for the large variance in the number of rooms present in NetHack maps, we show the average actions taken by each algorithm per number of map rooms in Figure~\ref{fig:comp1b}. The figure shows that as the number of rooms increases, the average actions of all algorithms rise linearly, but the occupancy map algorithm rises at a much slower rate than the greedy algorithm or BotHack. Thus, the occupancy map algorithm may be more efficient at handling more complicated or dense maps than the other two. The further variance that still exists in these results is due to the other randomness-induced properties of NetHack maps such as room location and corridor structure.}

\subsection*{Non-exhaustive approaches}

\begin{figure}
\centering
	\includegraphics[width=1\linewidth]{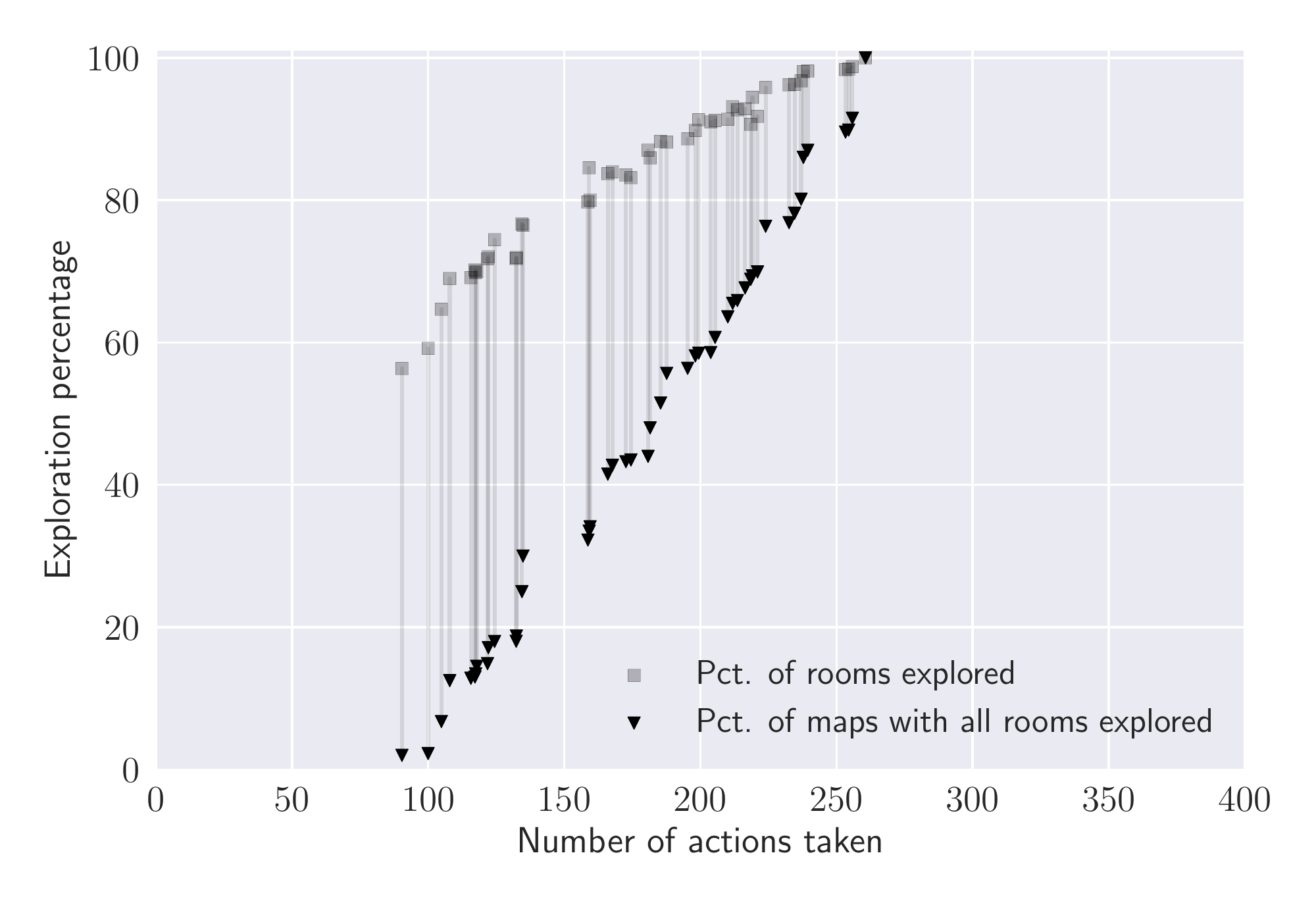}
	\caption{\small Occupancy map models with parameters that best minimize average actions per game and maximize percentage of rooms explored. Each symbol pair represents the average over 200 runs using a different combination of model parameters. The triangles show the result under the `exhaustive' metric and the corresponding squares show the total percentage of rooms explored.}
	\label{fig:occmap1}
\end{figure}

Exhaustive approaches are fine in certain circumstances, but it is often acceptable to occasionally leave one or two rooms on a map unexplored, especially when there is a cost to movement. Figure~\ref{fig:occmap1} gives the results for the best-performing non-exhaustive occupancy map models in terms of number of actions taken versus percentage of rooms explored. Each model represents an average over 200 runs using a unique combination of model parameters. (A grid search over the parameter space was performed -- the models shown lie on the upper-left curve of all models.)

As seen in the figure, there is a mostly linear progression in terms of the two metrics. The relationship between the `exhaustive' metric and total percentage of explored rooms is also consistent, with both linearly increasing.

The figure also shows that by sacrificing at most 5\% of room discovery \new{(or about 20\% of exhaustive room discovery)} on average, the average number of actions taken can be decreased to 200, compared to the 255 average actions of the exhaustive (98\%) approach.

\subsubsection*{Parameters}
\begin{figure}[htpb]
	\centering
	\includegraphics[width=0.5\textwidth]{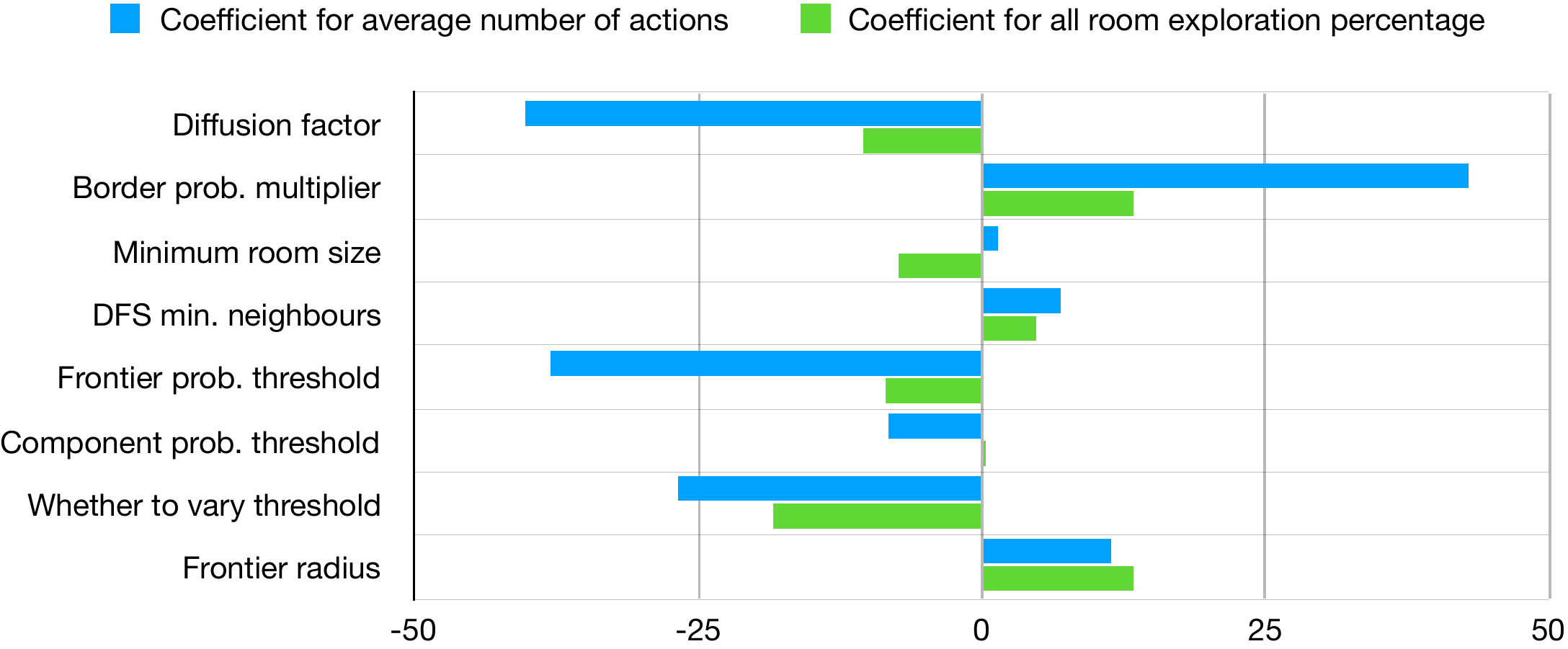}
	\caption{\small Linear regression coefficients for average number of actions and percentage of rooms explored with the occupancy map parameters as independent variables. Train/test data split is 70\%/30\% and dataset size is 5,823 (each datum being the result for a different combination of parameters).}
	\label{fig:occmap2}
\end{figure}

To determine the importance of the various parameters of the occupancy map algorithm, a linear regression was performed. Parameter coefficients for average actions and percentage of rooms explored under the `exhaustive' metric are shown in Figure~\ref{fig:occmap2}.  R-squared values for the regression were 0.781/0.521 (for average actions and room exploration) on test data. Running a random forest regressor on the same data gave the same general importances for each parameter with more confident r-squared values of 0.913/0.819, but those importances are not presented here due to lack of indication of the correlation direction.

The coefficients indicate that parameters directly associated with probabilities in the occupancy map are most effective with respect to average actions and percentage of rooms explored. These parameters include the diffusion factor (how much to diffuse to neighbours), border probability multiplier (how low the border cells should start at), both probability thresholds (at what probability to ignore frontiers/components), and whether to vary the threshold as more of the map is explored. The border diffusion is probably important due to the small (80x20) map size; on larger maps, it is less likely that this parameter would have such an impact.

The specific parameter values that led to the fastest performing exhaustive exploration model (presented in Figure~\ref{fig:comp1}) were as follows: diffusion factor of 0.65, distance importance of 1, border multiplier of 0.35, minimum room size of 5, DFS min. neighbours of 7, frontier probability threshold of 0.35, component threshold of 0.45, vary threshold set to false, and frontier radius of 2. \new{Distance factor (importance of distance in component evaluation) is kept constant to 1 in all experiments since it was shown to perform the best in early trials.}

\subsection*{Secret rooms}

\subsubsection*{Greedy algorithm for secret rooms}

\begin{figure}
\centering
	\includegraphics[width=0.95\linewidth]{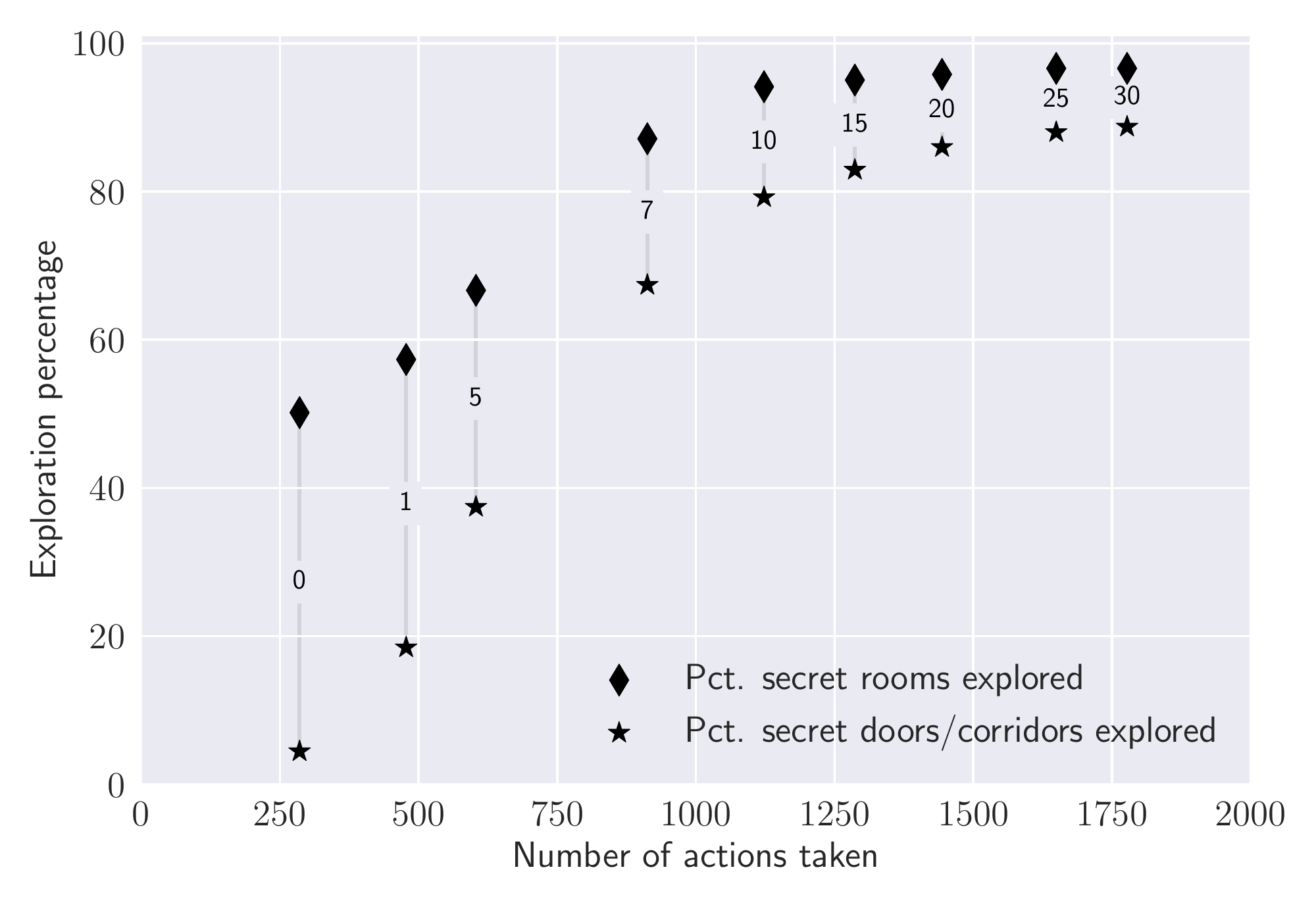}
	\caption{\small Greedy algorithm with support for secret door/corridor detection with labeled values showing number of searches per wall parameter. The average percentage of secret rooms found is represented by diamonds while average percentage of secret doors/corridors found is represented by stars.}
	\label{fig:secgreedy1}
\end{figure}

Figure~\ref{fig:secgreedy1} shows the results for the greedy algorithm with support for secret detection in terms of average actions versus exploration. Each symbol pair represents a different setting for the number of searches per wall parameter (the number of times each wall/dead-end corridor will be searched). Both the average percentage of secret rooms found and average percentage of secret doors/corridors found are displayed.

As expected, both metrics increase as the number of searches per wall increases, plateauing at around 95\% discovery of secret rooms and 90\% of secret doors/corridors at around 1750 average actions per game. As mentioned earlier, the algorithm will only search for secret corridors in dead-ends, so the 10\% of hidden spots not found at 30 searches per wall is most probably from secret corridors occurring (rarely) in other locations.

Another observation is that when the number of searches per wall is set to 0, the algorithm is reduced to the regular greedy algorithm, with almost no secret doors/corridors being found (no searching is performed, \new{but it is possible to accidentally discover a hidden spot by moving next to it}). The approximately 50\% score for the secret rooms metric is due to the fact that, in that percentage of runs, there were no secret rooms at all, thus giving 100\% exploration as mentioned in the metrics discussion.

\subsubsection*{Occupancy maps for secret rooms}

\begin{figure}
\centering
	\includegraphics[width=0.95\linewidth]{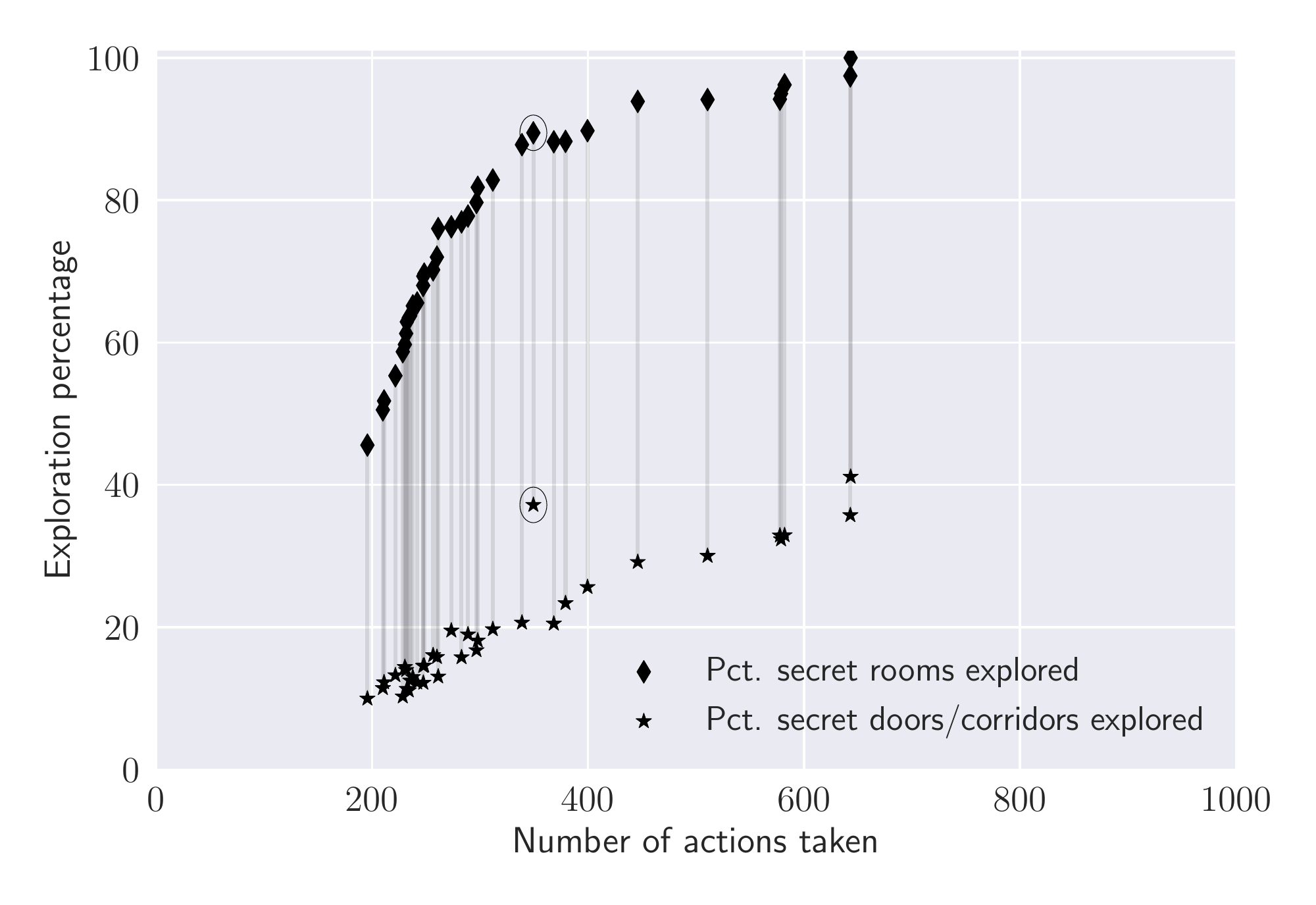}
	\caption{\small Occupancy map models with support for detecting secret doors/corridors, with parameters that best minimize average actions per game and maximize exploration of secret rooms. \new{The result for BotHack is shown, encircled.} Diamonds represent the percentage of secret rooms explored, while the corresponding stars represent the percentage of secret doors/corridors explored.}
	\label{fig:secmap1}
\end{figure}

Figure~\ref{fig:secmap1} gives the results for the best-performing secret-detecting occupancy map models in terms of number of actions taken versus secret area exploration. Each model represents an average over 200 runs using a unique combination of model parameters. (A grid search over the parameter space was performed; the models shown lie on the upper-left curve of all models.) \new{The result for BotHack is also shown.}

Results here are much better than the greedy algorithm, with approximately 90\% secret room exploration at under 500 actions. The discrepancy between this result and the greedy algorithm (over 1100 actions for 90\%) is because the occupancy map model has global knowledge of the map and can target particular walls for searching, in contrast to the greedy algorithm which searches almost every wall.

This global knowledge also explains the much lower percentage of secret doors/corridors discovered (about 30\% for the model exploring 90\% of secret rooms) compared to the greedy algorithm (80\% for the model exploring the same percentage of secret rooms). This result is expected since exploration of secret doors/corridors only weakly correlates with secret room exploration (only a few secret doors/corridors will actually lead to otherwise inaccessible rooms).

\new{Meanwhile, the occupancy map algorithm performs about the same as BotHack in terms of secret room discovery. The advantage with our algorithm is that it can be parameterized to do more or less exhaustive searching for secret doors, depending on the number of actions that can be spent.}

\new{One interesting result is the high amount of secret doors/corridors found by BotHack compared with our algorithm at the same number of actions/secret room exploration (about a 20\% difference). We speculate that this difference is because BotHack not only searches at locations likely to lead to secret rooms, but also searches to find `shortcuts' back to visited space, decreasing travel time between frontiers. However, this behaviour also increases the number of actions, and therefore does not translate into a lower number of actions than our algorithm, which finds the same percentage of secret rooms but less secret door/corridor shortcuts.}

\begin{figure}[htpb]
	\centering
	\includegraphics[width=0.5\textwidth]{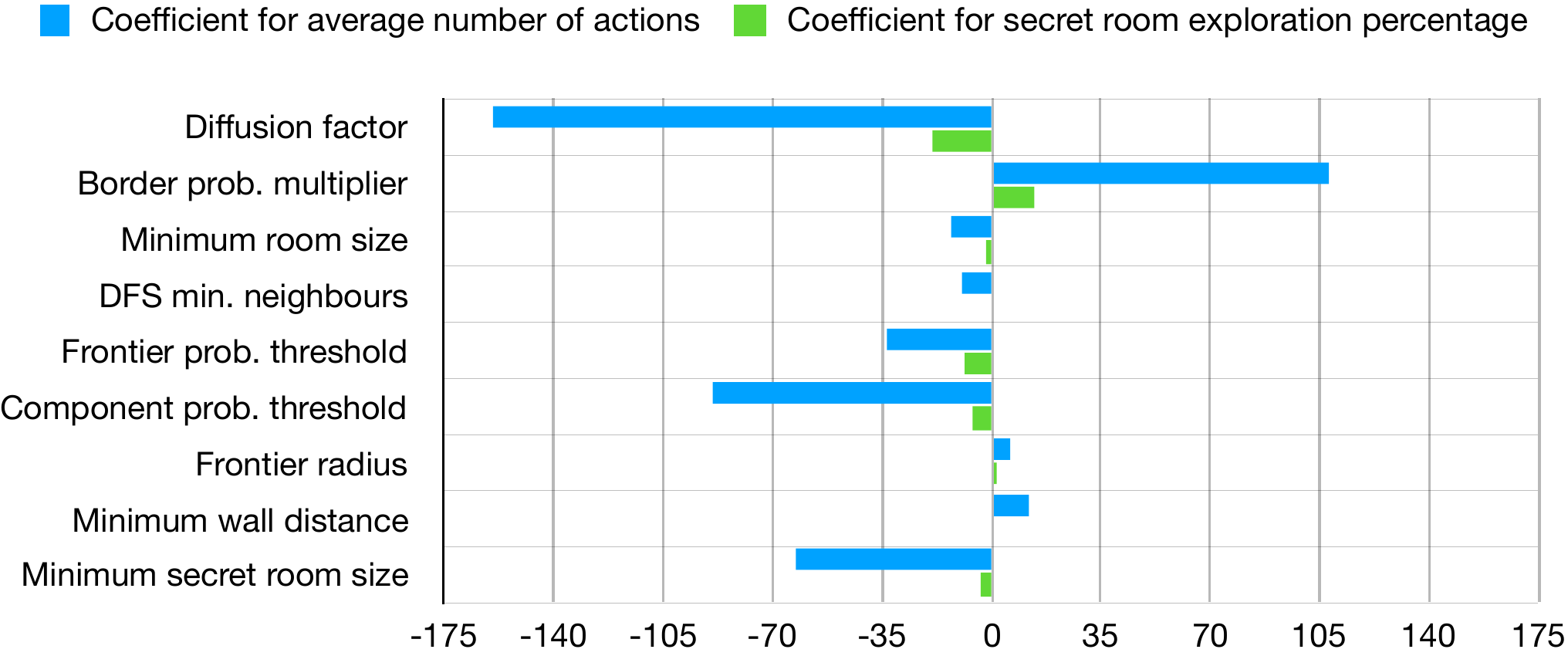}
	\caption{\small Linear regression coefficients for average number of actions and percentage of secret rooms explored with the secret-detecting occupancy map parameters as independent variables. Train/test data split is 70\%/30\% and dataset size is 1,792 (each datum being the result for a different combination of parameters).}
	\label{fig:secmap2}
\end{figure}

Importances of the parameters for the secret-detecting occupancy map algorithm are shown in Figure~\ref{fig:secmap2}. These importances were calculated by performing a \changed{linear regression} on the model results. R-squared values for the regression were \changed{0.742/0.809} (for average actions and secret room exploration) on test data. \new{We here keep the distance factor (importance of distance in component evaluation) to 1, wall distance factor to 1, vary probability threshold to False, and number of searches and maximum searches per wall to 10, to reduce the grid search computation time.}

The importances show that the probability-related parameters (diffusion factor, border multiplier and probability thresholds) continue to have a large impact on the average actions metric, as well as on the secret room exploration metric. \new{The minimum secret room size also has a large impact, greater than the impact of the minimum regular room size. This result suggests that components that can contain secret rooms are on average larger than regular components. These components without nearby frontiers could require more `padding' space for corridors that connect a secret room to observed space. Regular components, conversely, have frontiers that can be located much more closely to an unvisited room and thus require less padding space.}

\section{Conclusions and Future Work}
\label{sec:conclusions}


Automated exploration is an interesting and surprisingly complex task.  In
strategy or roguelike games, the tedium of repetitive movement during exploration is a concern
for players, and offering efficient automation can be helpful.  Exploration is also a significant
sub-problem in developing more fully automated, learning AI, and techniques which can algorithmically
solve exploration can be useful in allowing further automation to focus on applying AI to
higher level strategy rather than basic movement concerns.

In this work we detailed an algorithm for efficient exploration of an initially unknown
environment. Inspired by the occupancy map algorithm by Dami\'{a}n Isla for tracking a moving target,
we built an occupancy map approach to select frontiers to visit when performing exploration of
interesting areas of a map, while also considering complete coverage.  Our design notably improves over a
more straightforward, greedy design, particularly in the presence of secret areas, where exploration cost
versus benefit is especially important.

Our further work on the occupancy map algorithm aims at increasing efficiency in 
exploration. In particular, a `local' diffusion of probabilities (within a radius of the player position)
instead of the current global diffusion may prove fruitful to explore. Further verification of the
algorithm on other video games with different map configurations would also be interesting.

\section*{Acknowledgements}
This work supported by NSERC grant 249902.

\vspace*{0mm}

\bibliographystyle{IEEEtran}
\bibliography{bib}


\begin{IEEEbiographynophoto}{Jonathan Campbell}
 holds a B.A. and M.Sc. in Computer Science from McGill University, Montr\'eal, Canada. His research interests include artificial intelligence in games and applications of reinforcement learning.
\end{IEEEbiographynophoto}

\begin{IEEEbiography}[{\includegraphics[width=1in,height=1.25in,clip,keepaspectratio]{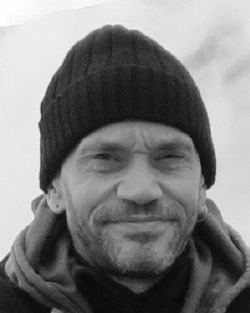}}]{Clark Verbrugge}
is an associate professor at the School of Computer Science
of McGill University in Montr\'eal, Canada. He is a member of both the \emph{Sable} 
and \emph{GR@M} groups at McGill, and does research in game formalization and analysis as well as
compiler and runtime optimization. He holds a Ph.D. in Computer Science from McGill University
and a B.A.(Hon.) in Film Studies from Queen's University.
\end{IEEEbiography}

\end{document}